%% file: arxiv.tex
\documentclass[10pt,twocolumn,letterpaper]{article}
\usepackage[accsupp]{axessibility}

\usepackage{cvpr}              % To produce the CAMERA-READY version
\input{preamble}

\definecolor{cvprblue}{rgb}{0.21,0.49,0.74}
\usepackage[pagebackref,breaklinks,colorlinks,citecolor=cvprblue]{hyperref}

\usepackage{hyperref}       % hyperlinks
\usepackage{url}            % simple URL typesetting
\usepackage{booktabs}       % professional-quality tables
\usepackage{amsfonts}       % blackboard math symbols
\usepackage{nicefrac}       % compact symbols for 1/2, etc.
\usepackage{microtype}      % microtypography
\usepackage{multirow}
\usepackage{graphicx}
\usepackage{colortbl}
\usepackage{caption}
\usepackage{amsmath}

\graphicspath{ {./Figure/} }

\makeatletter
\def\thickhline{%
\noalign{\ifnum0=`}\fi\hrule \@height \thickarrayrulewidth \futurelet
\reserved@a\@xthickhline}
\def\@xthickhline{\ifx\reserved@a\thickhline
            \vskip\doublerulesep
            \vskip-\thickarrayrulewidth
            \fi
    \ifnum0=`{\fi}}
\makeatother

\newlength{\thickarrayrulewidth}
\def\paperID{3038} % *** Enter the Paper ID here
\def\confName{CVPR}
\def\confYear{2024}

\title{SAFDNet: A Simple and Effective Network for Fully Sparse 3D Object Detection}

\author{Gang Zhang$^{1}$ \quad Junnan Chen$^{2}$ \quad Guohuan Gao$^{3}$ \quad Jianmin Li$^{1}$ \quad Si Liu$^{4}$ \quad Xiaolin Hu$^{1,5,6}\thanks{Corresponding Author}$\\
$^{1}$Department of Computer Science and Technology, Institute for AI, BNRist, Tsinghua University \\$^{2}$Huazhong University of Science and Technology\quad $^{3}$Beijing Institute of Technology \\
$^{4}$Institute of Artificial Intelligence, Beihang University\\
$^{5}$Tsinghua Laboratory of Brain and Intelligence (THBI), \\ IDG/McGovern Institute for Brain Research, Tsinghua University\\
$^{6}$Chinese Institute for Brain Research (CIBR), Beijing 100010, China \\
{\tt\small zhang-g19@mails.tsinghua.edu.cn, chen\_jn@hust.edu.cn, gaoguohuan@bit.edu.cn}, \\
{\tt\small liusi@buaa.edu.cn, lijianmin@mail.tsinghua.edu.cn, xlhu@tsinghua.edu.cn}
}
\begin{document}

\maketitle

\begin{abstract}
LiDAR-based 3D object detection plays an essential role in autonomous driving. Existing high-performing 3D object detectors usually build dense feature maps in the backbone network and prediction head. However, the computational costs introduced by the dense feature maps grow quadratically as the perception range increases, making these models hard to scale up to long-range detection. Some recent works have attempted to construct fully sparse detectors to solve this issue; nevertheless, the resulting models either rely on a complex multi-stage pipeline or exhibit inferior performance. In this work, we propose a fully sparse adaptive feature diffusion network (SAFDNet) for LiDAR-based 3D object detection. In SAFDNet, an adaptive feature diffusion strategy is designed to address the center feature missing problem. We conducted extensive experiments on Waymo Open, nuScenes, and Argoverse2 datasets. SAFDNet performed slightly better than the previous SOTA on the first two datasets but much better on the last dataset, which features long-range detection, verifying the efficacy of SAFDNet in scenarios where long-range detection is required. Notably, on Argoverse2, SAFDNet surpassed the previous best hybrid detector HEDNet by 2.6\% mAP while being 2.1$\times$ faster, and yielded 2.1\% mAP gains over the previous best sparse detector FSDv2 while being 1.3$\times$ faster. The code will be available at~\href{https://github.com/zhanggang001/HEDNet}{https://github.com/zhanggang001/HEDNet}.

\end{abstract}

\section{Introduction}

\begin{figure}
    \centering
    \includegraphics*[width=0.44\textwidth]{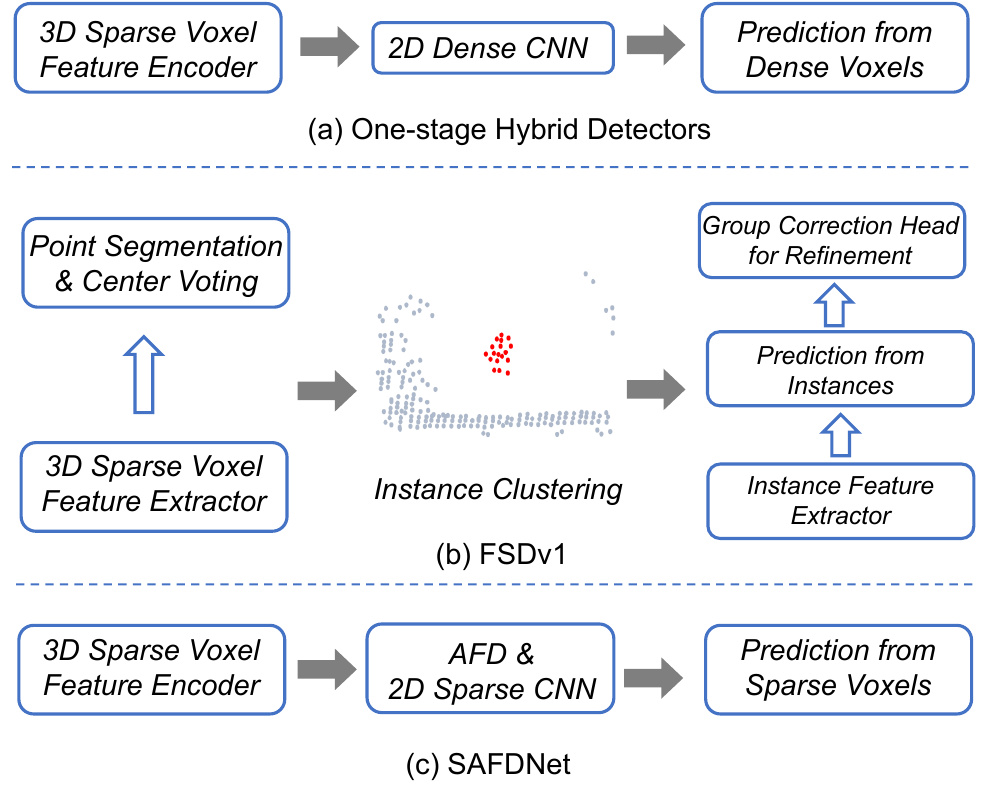}
    \vspace*{-3mm}
    \caption{Comparison among previous one-stage hybrid detectors, the fully sparse detector FSDv1, and our SAFDNet.}
    \label{comparison}
\end{figure}

LiDAR-based 3D object detection poses a significant challenge in computer vision and has received increasing attention for its potential applications in autonomous driving~\cite{chauffeurnet} and advanced robotics~\cite{zhu2017target}. Currently, most LiDAR-based 3D detectors~\cite{VoxelNet,CenterPoint,TransFusion,DSVT,HEDNet} convert sparse features into dense feature maps to facilitate further feature extraction and prediction, termed \textit{hybrid detectors} (see Figure~\ref{comparison}(a)). These methods demonstrate excellent performance on well-established benchmarks like nuScenes~\cite{nuScenes} and Waymo Open~\cite{Waymo}, primarily designed for a relatively short perception range (below 75 meters). However, scaling these methods to more practical long-range scenarios (exceeding 200 meters) becomes challenging because the computational costs associated with dense feature maps grow quadratically as the perception range increases~\cite{FSD}. Additionally, processing unoccupied areas is often unnecessary and might even hinder detection accuracy. Consequently, there's a growing interest among researchers in developing fully sparse detectors~\cite{SWFormer,FSD,FSDv2,VoxelNeXt}.

Constructing fully sparse detectors by removing dense feature maps from existing hybrid detectors is non-trivial, as these feature maps play a crucial role in these methods. Most hybrid detectors rely on features at object centers for predictions, considering them reliable representations of the entire object. These methods usually first employ sparse voxel encoders to efficiently extract features from non-empty voxels. Subsequently, they transform these sparse features into dense feature maps in a bird's eye view (BEV) and use convolutional neural networks (CNNs) to diffuse features towards object centers, creating center features. However, for fully sparse detector, in the absence of dense feature maps, the centers of large objects like vehicles and trucks often remain empty, leading to the \textit{center feature missing} problem~\cite{FSD,FSDv2}. Thus, learning an appropriate object representation becomes pivotal for building fully sparse detectors.

To tackle the center feature missing problem, FSDv1~\cite{FSD} proposes a multi-stage pipeline involving instance clustering (\mbox{Figure~\ref{comparison}(b)}). Specifically, it begins by segmenting raw point clouds into foreground and background, then conducts center voting for instance clustering. Subsequently, it extracts instance features from each cluster for initial predictions, which are refined by a group correction head. FSDv2~\cite{FSDv2} eliminates feature clustering in favor of a virtual voxelization module to reduce the inductive bias of handcrafted instance-level representations, yet it still relies on point segmentation and prediction refinement. The complex pipeline makes it challenging to deploy them in real-world scenarios due to numerous hyperparameters requiring tuning. In contrast, VoxelNeXt~\cite{VoxelNeXt} directly predicts objects based on the features nearest to object centers but exhibits inferior accuracy.

In this work, we introduce SAFDNet, a simple yet effective architecture tailored for fully sparse 3D object detection (Figure~\ref{comparison}(c)). Similar to hybrid detectors, SAFDNet initially employs a sparse voxel encoder to extract 3D sparse features, which are then transformed into 2D sparse BEV features. Subsequently, an adaptive feature diffusion (AFD) strategy is proposed to propagate features towards object centers, serving as the core component in SAFDNet for addressing the center feature missing problem. Unlike the uniform feature diffusion achieved by dense convolutional networks in hybrid detectors, our AFD selectively expands features within object bounding boxes to neighboring regions, dynamically adjusting the diffusion range according to voxel positions. As a result, SAFDNet can still leverage efficient calculations on sparse features. The expanded features are fed into the sparse detection head for predictions. Importantly, SAFDNet maintains most hyperparameters compatible with existing hybrid detectors, including those of the detection head, enabling easy adaptation to new scenarios.

We conducted extensive experiments on the challenging Waymo Open~\cite{Waymo}, nuScenes~\cite{nuScenes}, and Argoverse2~\cite{Argoverse} datasets to verify the effectiveness of our method. On the first two datasets for short-range detection, SAFDNet performed on par with the previous best hybrid detector HEDNet and was 2$\times$ faster than the previous best sparse detector FSDv2. On the Argoverse2 dataset for long-range detection, SAFDNet surpassed HEDNet by 2.6\% mAP while being 2.1$\times$ faster and outperformed FSDv2 by 2.1\% mAP while being 1.3$\times$ faster. These results demonstrate the efficacy of SAFDNet in scenarios that requires long-range detection.

\section{Related work}
\subsection{Dense detectors}
VoxelNet~\cite{VoxelNet} is the first to introduce dense convolutions for LiDAR-based 3D object detection, achieving competitive performance. However, directly applying dense convolutions to 3D voxel feature learning poses efficiency challenges due to their computational complexity. To address this limitation, pillar-based methods~\cite{PointPillar,PillarNet,PillarNeXt} utilize 2D dense convolutions on BEV dense feature maps instead, which improves computational efficiency but leads to inferior accuracy.

\subsection{Hybrid detectors}
Unlike dense detectors, hybrid detectors~\cite{SECOND,CenterPoint,CenterFormer,TransFusion,LargeKernel3D,SST,DSVT,HEDNet} incorporate both sparse and dense features. For instance, SECOND~\cite{SECOND}, a pioneering effort, employs a sparse CNN to extract 3D sparse voxel features and then transforms them into dense BEV feature maps for predictions. FocalsConv~\cite{FocalsConv} enhances the efficiency of sparse CNNs by adaptively expanding features through spatially learnable sparsity. CenterPoint~\cite{CenterPoint} introduces a center-based detection head, showcasing excellent performance in 3D object detection and tracking. Recent studies~\cite{CenterFormer,TransFusion,AFDetV2,HEDNet} have further enhanced CenterPoint from diverse perspectives. Additionally, another line of works~\cite{SST,SWFormer,votr,DSVT,EfficientTransformer} has explored transformers to capture long-range dependencies among spatial features. However, despite leveraging a sparse backbone, these methods face challenges when scaling to long-range scenarios, primarily due to their dependence on dense feature maps.

\begin{figure*}[t]
    \centering
    \includegraphics*[width=0.95\textwidth]{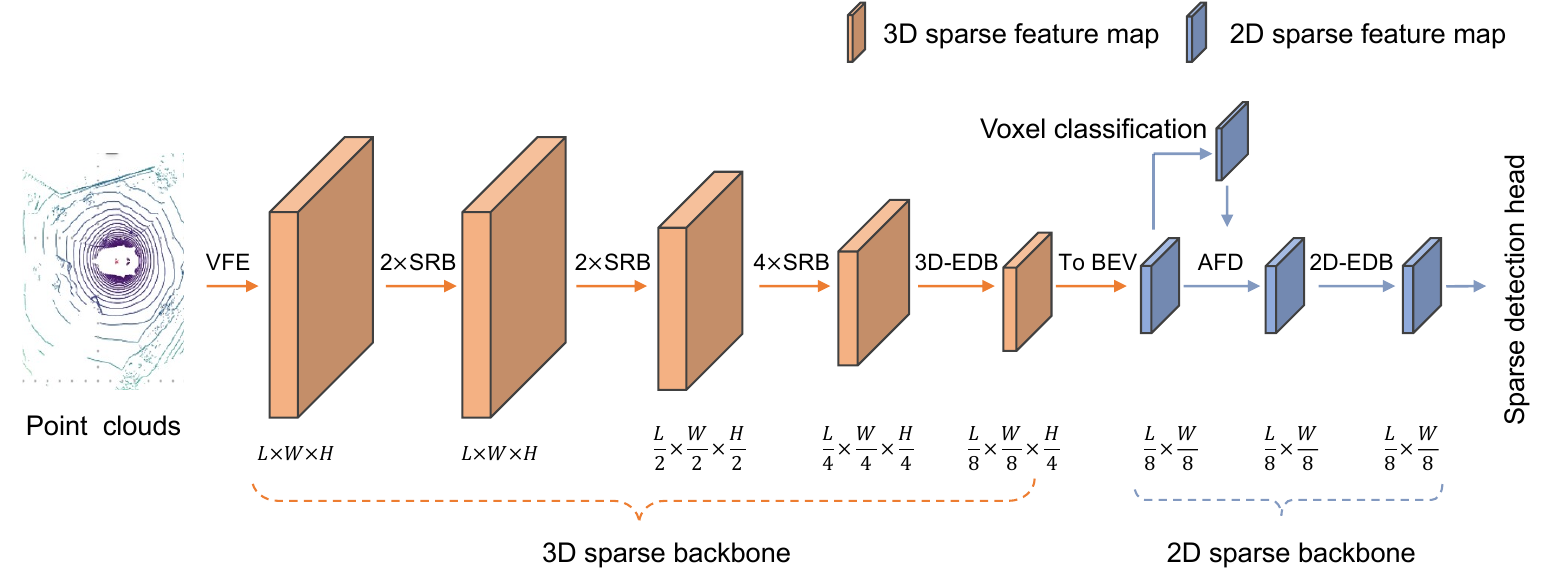}
    \vspace*{-4mm}
    \caption{Overall framework of SAFDNet. Taking the raw point clouds as input, SAFDNet extracts initial 3D sparse feature maps by the voxel feature encoder (VFE), and then it employs the 3D sparse backbone and the 2D sparse backbone to extract high-level sparse features for predictions in the sparse detection head. L, W and H denote length, width, and height of feature maps, respectively.}
    \label{framework}
\end{figure*}

\subsection{Sparse detectors}
Some early works~\cite{PointRCNN,FastPointRCNN,VoteNet} employ the PointNet series~\cite{PointNet,PointNet++} to extract sparse features from raw point clouds for predictions. Point R-CNN~\cite{PointRCNN} stands out as the pioneer in developing fully point-based detectors. VoteNet~\cite{VoteNet} introduces a center voting mechanism and generates proposals from the voted centers. Despite efforts to speed up full point-based methods, the time-consuming neighborhood searching remains impractical for large-scale point clouds. In contrast, FSDv1~\cite{FSD} segments raw point clouds into foreground and background, and then clusters the foreground points to represent individual objects. Then, it uses a PointNet-style~\cite{PointNet} network to extract features from each cluster for initial coarse predictions, refined by a group correction head. FSDv2~\cite{FSDv2} replaces the instance clustering with a virtual voxelization module, aiming to eliminate the inductive bias of handcrafted instance-level representations. Yet, it still requires point segmentation and prediction refinement. The complex pipeline demands tuning numerous hyperparameters for deployment in real-world scenarios. In contrast, SWFormer~\cite{SWFormer} presents a fully transformer-based architecture for 3D object detection. And the more recent VoxelNeXt~\cite{VoxelNeXt} streamlines the fully sparse architecture with a purely voxel-based design, localizing objects by the features nearest to their centers. Despite their notable efficiency, both SWFormer and VoxelNeXt exhibit inferior accuracy compared to hybrid detectors. % We reformulate the classification task to identify the voxels closest to object centers, enabling the regression task to predict precise bounding boxes using these closest voxel features.

\section{SAFDNet}

\subsection{Background}

\paragraph*{Sparse convolutions.}
Existing LiDAR-based 3D object detectors commonly leverage sparse convolutions for data processing to enhance computational efficiency. There are primarily two types of sparse convolutions used: \textit{submanifold sparse convolution}~\cite{submanifold_sparse}, which maintains feature sparsity between input and output feature maps, and \textit{regular sparse convolution}~\cite{regular_sparse}, which increases the density of the feature map by expanding features into neighboring regions. Since regular sparse convolution decreases feature sparsity dramatically, it is often merely used to down-sample feature maps in existing methods~\cite{SECOND,TransFusion,CenterPoint,HEDNet}.

\vspace*{-4mm}
\paragraph*{Sparse residual block (SRB).}
Most voxel-based methods~\cite{SECOND,CenterPoint,HEDNet} adopt sparse CNNs to extract features. These CNNs typically comprise a series of sparse residual blocks, where each block contains two submanifold sparse convolutions and a skip connection linking its input and output.

\begin{figure}[t]
    \centering
    \includegraphics*[width=0.47\textwidth]{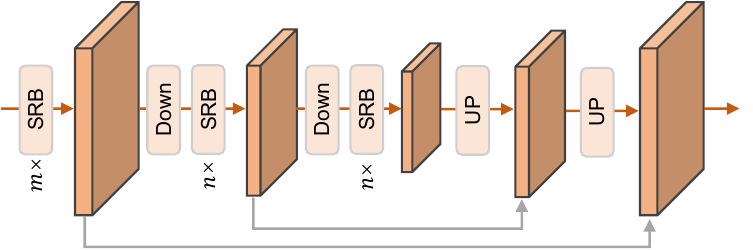}
    \vspace*{-1mm}
    \caption{Sparse encoder-decoder block. It adopts regular sparse convolution with stride 2 to down-sample feature maps and uses sparse inverse convolution~\cite{SPConv} to up-sample feature maps.}\label{EDB}
\end{figure}

\vspace*{-4mm}
\paragraph*{Sparse encoder-decoder block (EDB).} Since submanifold sparse convolutions preserve feature sparsity from input to output, they may impede information exchange among spatially distant features. As a result, merely stacking SRBs can result in a receptive field with limited size. HEDNet~\cite{HEDNet} addresses this by incorporating sparse encoder-decoder blocks that capture long-range dependencies among features while maintaining computational efficiency. Figure~\ref{EDB} illustrate a general structure of EDB. It decreases the spatial distance among distant features via feature down-sampling and subsequently restores lost details through multi-scale feature fusion. By applying 3D and 2D submanifold sparse convolutions to construct the SRB, we can obtain 3D-EDB and 2D-EDB, respectively.

\subsection{Overall architecture}
Figure~\ref{framework} presents an overview of the proposed SAFDNet. SAFDNet shares a similar pipeline to existing hybrid detectors~\cite{SECOND,CenterPoint}. It comprises three parts: a 3D sparse backbone, a 2D sparse backbone, and a sparse detection head.

\vspace*{-4mm}
\paragraph*{3D sparse backbone.} Taking raw point clouds as input, the 3D sparse backbone initially extracts sparse feature maps using a voxel feature encoder (VFE) and progressively down-samples them to extract high-level features. At the end of the backbone, it incorporates a 3D-EDB to facilitate information exchange among distant features. Subsequently, the 3D sparse features are compressed into 2D sparse BEV features. This compression is implemented by using two regular sparse convolutions with stride 2 to down-sample features along the Z-axis and then aggregating features of voxels sharing the same coordinates in BEV.

\vspace*{-4mm}
\paragraph*{2D sparse backbone.} Taking the BEV sparse features as input, the 2D sparse backbone begins by performing voxel classification on each voxel to determine whether the geometric center of each voxel falls within an object bounding box of a specific category or belongs to the background. Subsequently, an adaptive feature diffusion (AFD) operation, in conjunction with a 2D-EDB, is employed to propagate voxel features towards object centers.

\vspace*{-4mm}
\paragraph*{Sparse detection head.} Since most high-performing hybrid detectors utilize the center-based head introduced by CenterPoint~\cite{CenterPoint} or TransFusion-L~\cite{TransFusion}, we adopt a similar head in SAFDNet. However, as the designs of these detection heads are tailored for dense feature maps, we have made some adjustments to accommodate sparse features. For more details, please refer to Section~\ref{detection_head}.

\begin{table}
    \centering
    \scalebox{0.9}{
        \setlength{\tabcolsep}{1.0mm}{}
        \begin{tabular}{c|c|cccc}
        \toprule
        Method & Type & mAPH & Vehicle & Pedestrian & Cyclist  \\
        \midrule
        HEDNet & Hybrid & 73.2 & 72.1 & 72.0 & 75.6 \\
        Nearest & Sparse & 71.5 & 68.9 & 70.9 & 74.7 \\
        \bottomrule
        \end{tabular}
    }
    \vspace*{-2mm}
    \caption{Preliminary experiments on the Waymo Open dataset. The second model shares a similar structure to HEDNet but replaces all dense convolutions with submanifold sparse convolutions. The overall accuracy mAPH and the per-category APH are presented.}
    \label{motivation_exps}
    % \vspace*{-1mm}
 \end{table}

\subsection{Adaptive feature diffusion}
Existing hybrid detectors typically decompose 3D object detection into classification and regression tasks. The classification task aims to locate voxels at object centers for each category, while the regression task predicts precise bounding boxes based on these center features. Given that LiDAR point clouds reside on the surfaces of objects, to construct fully sparse detectors by simply removing dense feature maps would result in the center feature missing problem. A straightforward solution is to make predictions based on the features nearest to object centers. Specifically, we reformulate the classification task into identifying the voxels closest to object centers, enabling the regression task to predict precise bounding boxes using these closest voxel features. Our experiments, detailed in Table~\ref{motivation_exps}, demonstrate that such a sparse model (the bottom row) performed worse than the hybrid detector HEDNet. This discrepancy is particularly notable on larger vehicles, which are more severely affected by the center feature missing problem. These findings indicate that center features indeed provide better object representations than their nearest counterparts.

\begin{figure}[t]
    \centering
    \includegraphics*[width=0.47\textwidth]{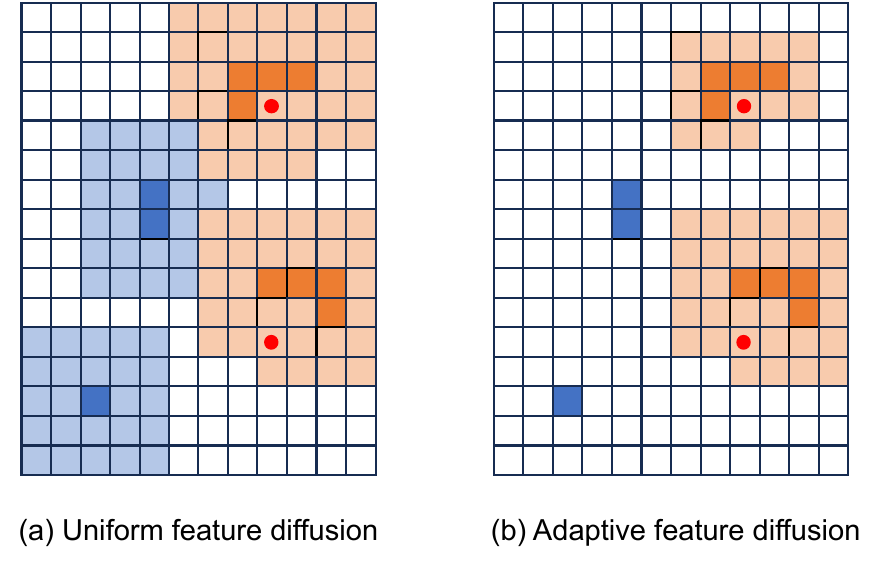}
    \vspace*{-4mm}
    \caption{Illustration of uniform and adaptive feature diffusion. The red points denote object centers. The voxels with centers falling within object bounding boxes are indicated in dark orange, while those outside are in dark blue. The expanded features are indicated in light orange or light blue. Empty voxels are indicated in white.}
    \label{feature_diffusion}
\end{figure}

\vspace*{-1mm}
\paragraph*{Uniform feature diffusion (UFD).} Is it possible for detectors to extract features nearer or at object centers while maintaining feature sparsity as much as possible? An intuitive idea is to diffuse sparse features to neighboring voxels rather than all voxels like hybrid detectors. Figure~\ref{feature_diffusion} (a) depicts a uniform feature diffusion strategy, where input voxel features are expanded to a $K\times K$ neighborhood, with $K$ set to 5 as an example. There are two possible implementations:

\begin{enumerate}[label=\alph*.]
\item Parameter-based (PB) UFD: employing a regular sparse convolution with kernel size $K\times K$ to spread features, in conjunction with a 2D-EDB for further transformation.
\item Parameter-free (PF) UFD: initializing zero features in neighboring regions and then incorporating a 2D-EDB to spread features progressively.
\end{enumerate}

\vspace*{-4mm}
\paragraph*{Adaptive feature diffusion (AFD).}
Through our analysis of the sparse voxels output by the 3D sparse backbone, we observed that: (a) fewer than 10\% of the voxels fall within the bounding boxes of objects; (b) smaller objects often have voxel features near or at their centers. This indicates potential redundancy in uniformly expanding features into neighborhoods of the same size, particularly for voxels within the bounding boxes of small objects and those belonging to the background. Hence, we propose an adaptive feature diffusion strategy according to voxel positions, as depicted in Figure~\ref{feature_diffusion} (b). \textit{\textbf{The idea}} is to assign a larger diffusion range to voxels within the bounding boxes of large objects to bring features nearer to object centers, while assigning a smaller range to voxels within the bounding boxes of small objects or the background to preserve feature sparsity. Implementing this idea necessitates \textit{\textbf{voxel classification}} to determine whether a voxel's center is within the bounding box of an object of a specific category or belongs to the background.

\begin{table*}[t]
    \begin{center}
    \scalebox{0.9}{
       \setlength{\tabcolsep}{1.8mm}{}
       \hspace{-4mm}
       \begin{tabular}{l|c|c|c|c|c|c|c|c}
       \thickhline
       \multirow{2}{*}{Method} & \multicolumn{2}{c|}{mAP/mAPH} & \multicolumn{2}{c|}{Vehicle AP/APH} &
       \multicolumn{2}{c|}{Pedestrian AP/APH} & \multicolumn{2}{c}{Cyclist AP/APH} \\
       & L1 & L2 & L1 & L2 & L1 & L2 & L1 & L2 \\
       \midrule
       \multicolumn{9}{c}{\textit{Results on the validation data set}} \\
       \midrule
       SECOND~\cite{SECOND} & 67.2/63.1 & 61.0/57.2 & 72.3/71.7 & 63.9/63.3 & 68.7/58.2 & 60.7/51.3 & 60.6/59.3 & 58.3/57.0\\
       PointPillar~\cite{PointPillar} & 69.0/63.5 & 62.8/57.8 & 72.1/71.5 & 63.6/63.1 & 70.6/56.7 & 62.8/50.3 & 64.4/62.3 & 61.9/59.9 \\
    %    Lidar-RCNN~\cite{LIDAR-RCNN} & 71.9/67.0 & 65.8/61.3 & 76.0/75.5 & 68.3/67.9 & 71.2/58.7 & 63.1/51.7 & 68.6/66.9 & 66.1/64.4 \\
       Part-A2-Net~\cite{PartA2Net} & 73.6/70.3 & 66.9/63.8 & 77.1/76.5 & 68.5/68.0 & 75.2/66.9 & 66.2/58.6 & 68.6/67.4 & 66.1/64.9 \\
       SST~\cite{SST} & 74.5/71.0 & 67.8/64.6 & 74.2/73.8 & 65.5/65.1 & 78.7/69.6 & 70.0/61.7 & 70.7/69.6 & 68.0/66.9 \\
       CenterPoint~\cite{CenterPoint} & 74.4/71.7 &68.2/65.8 & 74.2/73.6 & 66.2/65.7 & 76.6/70.5 & 68.8/63.2 & 72.3/71.1 & 69.7/68.5 \\
       PV-RCNN~\cite{PV-RCNN} & 76.2/73.6 & 69.6/67.2 & 78.0/77.5 & 69.4/69.0 & 79.2/73.0 & 70.4/64.7 & 71.5/70.3 & 69.0/67.8 \\
       CenterPoint~\cite{CenterPoint} & 75.9/73.5 & 69.8/67.6 & 76.6/76.0 & 68.9/68.4 & 79.0/73.4 & 71.0/65.8 & 72.1/71.0 & 69.5/68.5 \\
    %    OcTr~\cite{OcTr} & 77.2/74.2 & 70.7/68.2 & 78.1/77.6 & 69.8/69.3 & 80.8/74.4 & 72.5/66.5 & 72.6/71.5 & 69.9/68.9 \\
       PillarNet-34~\cite{PillarNet} & 77.3/74.6 & 71.0/68.5 & 79.1/78.6 & 70.9/70.5 & 80.6/74.0 & 72.3/66.2 & 72.3/71.2 & 69.7/68.7 \\
       AFDetV2~\cite{AFDetV2} & 77.2/74.8 & 71.0/68.8 & 77.6/77.1 & 69.7/69.2 & 80.2/74.6 & 72.2/67.0 & 73.7/72.7 & 71.0/70.1 \\
       CenterFormer~\cite{CenterFormer} & 75.6/73.2 & 71.4/69.1 & 75.0/74.4 & 69.9/69.4 & 78.0/72.4 & 73.1/67.7 & 73.8/72.7 & 71.3/70.2 \\
       LargeKernel3D\cite{LargeKernel} & -/- & -/- & 78.1/77.6 & 69.8/69.4 & -/- & -/- & -/- & -/- \\
       PV-RCNN++~\cite{PV-RCNN++} & 78.1/75.9 & 71.7/69.5 & 79.3/78.8 & 70.6/70.2 & 81.3/76.3 & 73.2/68.0 & 73.7/72.7 & 71.2/70.2 \\
       DSVT-Voxel~\cite{DSVT} & 80.3/78.2 & 74.0/72.1 & 79.7/79.3 & 71.4/71.0 & 83.7/78.9 & 76.1/71.5 & 77.5/76.5 & 74.6/73.7 \\
       HEDNet~\cite{HEDNet} & 81.4/79.4 & 75.3/73.4 & 81.1/80.6 & 73.2/72.7 & 84.4/80.0 & 76.8/72.6 & 78.7/77.7 & 75.8/74.9 \\
       \midrule
       SWFormer~\cite{SWFormer}$^\dag$ & -/- &  -/- & 77.8/77.3 & 69.2/68.8 & 80.9/72.7 & 72.5/64.9 & -/- & -/- \\
    %    FlatFormer~\cite{FlatFormer}$^\dag$ & 79.3/77.1 & 72.7/70.5 & 78.6/78.1 & 69.8/69.4 & 82.9/77.5 & 74.3/69.3 & 76.6/75.6 & 73.9/72.8 \\
       VoxelNeXt~\cite{VoxelNeXt}$^\dag$ & 78.6/76.3 & 72.2/70.1 & 78.2/77.7 & 69.9/69.4 & 81.5/76.3 & 73.5/68.6 & 76.1/74.9 & 73.3/72.2 \\
       FSDv1~\cite{FSD}$^\dag$ & 79.6/77.4 & 72.9/70.8 & 79.2/78.8 & 70.5/70.1 & 82.6/77.3 & 73.9/69.1 & 77.1/76.0 & 74.4/73.3 \\
       FSDv2~\cite{FSDv2}$^\dag$ & 81.8/79.5 & 75.6/73.5 & 79.8/79.3 & 71.4/71.0 & 84.8/79.7 & 77.4/72.5 & 80.7/79.6 & 77.9/76.8 \\
       \rowcolor{cyan! 20} SAFDNet (ours)$^\dag$ & 81.8/79.9 & \textbf{75.7/73.9} & 80.6/80.1 & 72.7/72.3 & 84.7/80.4 & 77.3/73.1 & 80.0/79.0 & 77.2.76.2 \\
       \midrule
       \multicolumn{9}{c}{\textit{Results on the test data set}} \\
    %    \midrule
    %    \multirow{2}{*}{Method} & mAP/mAPH & mAP/mAPH & \multicolumn{2}{c|}{Vehicle AP/APH} &
    %    \multicolumn{2}{c|}{Pedestrian AP/APH} & \multicolumn{2}{c}{Cyclist AP/APH} \\
    %    & L1 & L2 & L1 & L2 & L1 & L2 & L1 & L2 \\
       \midrule
    %    PV-RCNN~\cite{PV-RCNN}$^\dag$ & 76.9/74.2 & 71.3/68.8 & 80.6/80.1 & 72.8/72.4 & 78.2/72.0 & 71.8/66.0 & 71.8/70.4 & 69.1/67.8 \\
    %    PV-RCNN++~\cite{PV-RCNN++}$^\dag$ & 78.0/75.7 & 72.4/70.2 & 81.6/81.2 & 73.9/73.5 & 80.4/75.0 & 74.1/69.0 & 71.9/70.8 & 69.3/68.2 \\
    %    AFDetV2~\cite{AFDetV2} & 77.6/75.2 & 72.2/70.3 & 80.5/80.0 & 73.0/72.6 & 79.8/74.3 & 73.7/68.6 & 72.4/71.2 & 69.8/69.7 \\
    %    HEDNet~\cite{HEDNet} & 82.2/80.2 & 76.9/75.0 & 84.2/83.8 & 77.0/76.6 & 84.1/79.7 & 78.3/74.0 & 78.2/77.0 & 75.4/74.3 \\
    %    \midrule
       FSDv1~\cite{FSD}$^\dag$ & 80.4/78.2 & 74.4/72.4 & 82.7/82.3 & 74.4/74.1 & 82.9/77.9 & 75.9/71.3 & 75.6/74.4 & 72.9/71.8 \\
       FSDv2~\cite{FSDv2}$^\dag$ & 81.1/79.0 & 75.4/73.3 & 82.4/82.0 & 74.4/74.0 & 83.8/78.9 & 77.4/72.8 & 77.1/76.0 & 74.3/73.2 \\
       \rowcolor{cyan! 20} SAFDNet (ours)$^\dag$ & 81.9/79.8 & \textbf{76.5/74.6} & 83.9/83.5 & 76.6/76.2 & 84.3/79.8 & 78.4/74.1 & 77.5/76.3 & 74.6/73.4 \\
       \thickhline
       \end{tabular}
    }
    \end{center}
    \vspace*{-5mm}
    \caption{Comparison with prior methods on the Waymo Open dataset. Metrics: mAP/mAPH (\%)$\uparrow$ for the overall results, and AP/APH (\%)$\uparrow$ for each category. $^\dag$ represents a fully sparse detector, the same as below. All models were trained under single-frame setting.}
    \label{waymo_results}
\end{table*}

\vspace*{-4mm}
\paragraph*{Training.} For the training process of voxel classification, we group object categories of similar size and perform binary classification for each group. Let $G$ denote the number of category groups and $N$ be the number of sparse voxels. For group $i$, the model predicts a vector $\mathbf{P}_i$ of length $N$. The corresponding training target $\mathbf{T}_i$ is defined as follows:
\vspace*{-1mm}
\begin{equation}
    \mathbf{T}_i^j = \left\{
        \begin{array}{lr}
            \kern-0.5em1, \ \text{if}\ (x_j, y_j)\ \text{is in object box of group}\ i &  \\
            \kern-0.5em0, \ \text{otherwise} &
        \end{array}
\right.
\vspace*{-2mm}
\end{equation}
\noindent where $j\kern-0.2em\in\kern-0.2em\{1, ..., N\}$, $(x_j, y_j)$ represents the coordinates of the corresponding voxel center, and the `object box' refers to the human-annotated bounding box of an object. The overall loss for voxel classification is defined by the equation:
\vspace*{-2mm}
\begin{equation}
    L_{\mathrm{AFD}} = \sum_{i=1}^G \text{sigmoid\_focal\_loss}(\mathbf{P}_i, \mathbf{T}_i)
    \vspace*{-2mm}
\end{equation}

\vspace*{-4mm}
\paragraph*{Inference.} Given $\mathbf{P}_i$,  a binary mask $\mathbf{M}_i$ that indicates voxels whether fall within any object bounding boxes of category group $i$ is calculated by threshold $t$, then the binary mask $\mathbf{R}_i$ that indicates the feature diffusion areas can be jointly decided by $\mathbf{M}_i$ and the diffusion kernel size $K_i\kern-0.2em\times\kern-0.2em K_i$. $K_i$ is defined as $\alpha\cdot S_i$, where $S_i$ is the average size of objects in category group $i$ and $\alpha$ controls the range of feature diffusion. The object size is defined as the maximum value between the length and width of the bounding box and is normalized by the voxel size. After that, the binary mask representing the overall feature diffusion areas can be calculated as $\mathbf{R}=\mathbf{R}_1\cup \mathbf{R}_2 \cup ... \cup \mathbf{R}_G$. For each voxel in the feature diffusion areas, if there are no features, we initialize it with zero features. Finally, a 2D-EDB is utilized for further feature transformation.

% \vspace*{-5mm}
% \paragraph*{Discussion.} The previous fully sparse detectors, FSDv1 and FSDv2, utilize point segmentation, center voting, and instance clustering (or virtual voxel mixer) in the raw point cloud space to tackle the center feature missing problem. The complex pipelines introduce numerous hyperparameters, making their application to new scenarios challenging. In contrast, our proposed AFD strategy offers a straightforward solution with only a few hyperparameters. Experiments in Section~\ref{ablation_exps} demonstrate the effectiveness of our method.

\begin{table*}[t]
    \begin{center}
    \scalebox{0.9}{
       \setlength{\tabcolsep}{2.0mm}{}
       \hspace{-4mm}
       \begin{tabular}{l|cc|cccccccccc}
       \thickhline
      Method & NDS & mAP & Car & Truck & Bus & T.L. & C.V. & Ped. & M.T. & Bike & T.C. & B.R.\\
      \midrule
      \multicolumn{13}{c}{\textit{Results on the validation data set}} \\

    %   CenterPoint~\cite{CenterPoint} & 66.5 & 59.2 & 84.9 & 57.4 & 70.7 & 38.1 & 16.9 & 85.1 & 59.0 & 42.0 & 69.8 & 68.3 \\
    %   TransFusion-L~\cite{TransFusion} & 70.1 & 65.5 & 86.9 & 60.8 & 73.1 & 43.4 & 25.2 & 87.5 & 72.9 & 57.3 & 77.2 & 70.3 \\
    %   HEDNet~\cite{HEDNet} & 71.4 & 66.7 & 87.7 & 60.6 & 77.8 & 50.7 & 28.9 & 87.1 & 74.3 & 56.8 & 76.3 & 66.9 \\
      \midrule
      VoxelNeXt~\cite{VoxelNeXt}$^\dag$ & 68.7 & 63.5 & 83.9 & 55.5 & 70.5 & 38.1 & 21.1 & 84.6 & 62.8 & 50.0 & 69.4 & 69.4 \\
      FSDv2~\cite{FSDv2}$^\dag$ & 70.4 & 64.7 & 84.4 & 57.3 & 75.9 & 44.1 & 28.5 & 86.9 & 69.5 & 57.4 & 72.9 & 73.6 \\
      \rowcolor{cyan! 20} SAFDNet (Ours)$^\dag$ & \textbf{71.0} & \textbf{66.3} & 87.6 & 60.8 & 78.0 & 43.5 & 26.6 & 87.8 & 75.5 & 58.0 & 75.0 & 69.7 \\
      \midrule
      \multicolumn{13}{c}{\textit{Results on the test data set}} \\
    %   \midrule
    %   Method & NDS & mAP & Car & Truck & Bus & T.L. & C.V. & Ped. & M.T. & Bike & T.C. & B.R.\\
      \midrule
      PointPillars~\cite{PointPillar} & 45.3 & 30.5 & 68.4 & 23.0 & 28.2 & 23.4 & 4.1 & 59.7 & 27.4 & 1.1 & 30.8 & 38.9\\
      3DSSD~\cite{3DSSD} & 56.4 & 42.6 & 81.2 & 47.2 & 61.4 & 30.5 & 12.6 & 70.2 & 36.0 & 8.6 & 31.1 & 47.9\\
    %   CBGS~\cite{CBGS} & 63.3 & 52.8 & 81.1 & 48.5 & 54.9 & 42.9 & 10.5 & 80.1 & 51.5 & 22.3 & 70.9 & 65.7\\
      CenterPoint~\cite{CenterPoint} & 65.5 & 58.0 & 84.6 & 51.0 & 60.2 & 53.2 & 17.5 & 83.4 & 53.7 & 28.7 & 76.7 & 70.9\\
    %   FCOS-LiDAR~\cite{FCOS3D} & 65.7 & 60.2 & 82.2 & 47.7 & 52.9 & 48.8 & 28.8 & 84.5 & 68.0 & 39.0 & 79.2 & 70.7 \\
      HotSpotNet~\cite{HotSpotNet} & 66.0 & 59.3 & 83.1 & 50.9 & 56.4 & 53.3 & 23.0 & 81.3 & 63.5 & 36.6 & 73.0 & 71.6\\
      CVCNET~\cite{CVCNet} & 66.6 & 58.2 & 82.6 & 49.5 & 59.4 & 51.1 & 16.2 & 83.0 & 61.8 & 38.8 & 69.7 & 69.7\\
      AFDetV2~\cite{AFDetV2} & 68.5 & 62.4 & 86.3 & 54.2 & 62.5 & 58.9 & 26.7 & 85.8 & 63.8 & 34.3 & 80.1 & 71.0 \\
      UVTR-L~\cite{UVTR} & 69.7 & 63.9 & 86.3 & 52.2 & 62.8 & 59.7 & 33.7 & 84.5 & 68.8 & 41.1 & 74.7 & 74.9\\
      VISTA~\cite{VISTA} & 69.8 & 63.0 & 84.4 & 55.1 & 63.7 & 54.2 & 25.1 & 82.8 & 70.0 & 45.4 & 78.5 & 71.4\\
      Focals Conv~\cite{FocalsConv} & 70.0 & 63.8 & 86.7 & 56.3 & 67.7 & 59.5 & 23.8 & 87.5 & 64.5 & 36.3 & 81.4 & 74.1\\
      TransFusion-L~\cite{TransFusion} & 70.2 & 65.5 & 86.2 & 56.7 & 66.3 & 58.8 & 28.2 & 86.1 & 68.3 & 44.2 & 82.0 & 78.2\\
      LargeKernel3D~\cite{LargeKernel3D} & 70.6 & 65.4 & 85.5 & 53.8 & 64.4 & 59.5 & 29.7 & 85.9 & 72.7 & 46.8 & 79.9 & 75.5\\
      LinK~\cite{LinK} & 71.0 & 66.3 & 86.1 & 55.7 & 65.7 & 62.1 & 30.9 & 85.8 & 73.5 & 47.5 & 80.4 & 75.5\\
      HEDNet~\cite{HEDNet} & 72.0 & 67.7 & 87.1 & 56.5 & 70.4 & 63.5 & 33.6 & 87.9 & 70.4 & 44.8 & 85.1 & 78.1 \\
      \midrule
      VoxelNeXt~\cite{VoxelNeXt}$^\dag$ & 70.0 & 64.5 & 84.6 & 53.0 & 64.7 & 55.8 & 28.7 & 85.8 & 73.2 & 45.7 & 79.0 & 74.6 \\
      FSDv2~\cite{FSDv2}$^\dag$ & 71.7 & 66.2 & 83.7 & 51.6 & 66.4 & 59.1 & 32.5 & 87.1 & 71.4 & 51.7 & 80.3 & 78.7 \\
      \rowcolor{cyan! 20} SAFDNet (Ours)$^\dag$ & \textbf{72.3} & \textbf{68.3} & 87.3 & 57.3 & 68.0 & 63.7 & 37.3 & 89.0 & 71.1 & 44.8 & 84.9 & 79.5 \\
       \hline
       \thickhline
       \end{tabular}}
    \end{center}
    \vspace*{-5mm}
    \caption{Comparison with prior methods on the nuScenes dataset. Metrics: NDS (\%)$\uparrow$ and mAP (\%)$\uparrow$ for the overall results, AP (\%)$\uparrow$ for each category. ‘T.L.’, ‘C.V.’, ‘Ped.’, ‘M.T.’, ‘T.C.’, and 'B.R.' denote trailer, construction vehicle, pedestrian, motor, traffic cone, and barrier.}
    \label{nuscenes_results}
\end{table*}

\subsection{Sparse detection head}\label{detection_head}
We adhere to most design principles of CenterPoint (Waymo Open and Argoverse2) and TransFusion-L (nuScenes), but make some adjustments to accommodate sparse features. Despite employing adaptive feature diffusion, covering all object centers remains challenging, particularly for extremely large objects. Therefore, we adopt the diffused voxel features nearest to object centers as object representations. For classification training, we calculate a Gaussian heatmap for each object based on the distance from each voxel center to the object center. We normalize the heatmap values by the maximum value for each object to avoid gradient vanishing. %For regression in CenterPoint, we compute the loss based on predictions from voxels features nearest to object centers.

\section{Experiments}

% \subsection*{Experimental settings}
% We conducted experiments on the three autonomous driving datasets including Waymo Open~\cite{Waymo}, nuScenes~\cite{nuScenes} and Argoverse2~\cite{Argoverse} to validate the effectiveness of our approach. The detection ranges of the three datasets are 75 meters, 54 meters and 200 meters, respectively. To build SAFDNet, we set the hyperparameters $m$, $n$ to 4, 2 for the 3D-EDB, and 8, 4 for the 2D-EDB. Additionally, the hyperparameters $t$ and $\alpha$ in AFD were set to 0.5 and 1.0. All experiments were conducted on 8 RTX 4090 GPUs with a total batch size of 16. To compare with previous state-of-the-art methods, we trained SAFDNet for 24 epochs, 20 epochs, and 24 epochs on the Waymo Open, nuScenes, and Argoverse2 datasets, respectively. For ablation experiments in Section~\ref{ablation_exps}, we trained models for 12 epochs and 6 epochs on the Waymo Open and Argoverse2 datasets, respectively. Please refer to the \textbf{Appendix} for more details.

\subsection{Datasets and metrics}

We conducted experiments on the popular Waymo Open~\cite{Waymo}, nuScenes~\cite{nuScenes}, and Argoverse2~\cite{Argoverse} datasets to validate the effectiveness of our approach. The detection ranges of the three datasets are 75, 54 and 200 meters, respectively. For object detection on the \textit{Waymo Open} dataset, evaluation metrics include mean average precision (mAP) and mAP weighted by heading accuracy (mAPH). Both are further broken down into two difficulty levels: L1 for objects with more than five LiDAR points and L2 for objects with at least one LiDAR point. For object detection on the \textit{nuScenes} dataset, evaluation metrics include mAP and the nuScenes detection score (NDS). mAP is calculated by averaging over the distance thresholds of 0.5m, 1m, 2m, and 4m across all categories. NDS is a weighted average of mAP and five other true positive metrics that measure translation, scaling, orientation, velocity, and attribute errors. For object detection on the \textit{Argoverse2} dataset, mAP is adopted as the evaluation metric.

\subsection{Implementations details}
We implemented our method based on the open-source OpenPCDet~\cite{openpcdet}. To build SAFDNet, we set the hyperparameters $m$, $n$ to 4, 2 for the 3D-EDB, and 8, 4 for the 2D-EDB. The hyperparameters $t$ and $\alpha$ in AFD were set to 0.4 and 1.0. All experiments were conducted on 8 RTX 4090 GPUs with a total batch size of 16. To compare with previous state-of-the-art methods, we trained SAFDNet for 24 epochs, 20 epochs, and 24 epochs on the Waymo Open, nuScenes, and Argoverse2 datasets, respectively. For ablation experiments in Section~\ref{ablation_exps}, we trained models for 12 epochs and 6 epochs on the Waymo Open and Argoverse2 datasets, respectively. Please refer to the \textbf{Appendix A} for more details.

% \vspace*{-4mm}
% \paragraph*{Waymo Open.} We set the voxel size to (0.08m, 0.08m, 0.15m) and the detection range to [-75.2m, 75.2m] in X and Y axes, and [-2m, 4m] in Z axis. We trained SAFDNet for 24 epochs on the entire training dataset (\textit{single-frame}) to compare with previous methods. For the ablation experiments in Section~\ref{ablation_exps}, we trained all models for 12 epochs. All models were trained with a batch size of 16 on 8 RTX 4090 GPUs. The other training settings strictly followed HEDNet~\cite{HEDNet}.

% \vspace*{-4mm}
% \paragraph*{nuScenes} We set the voxel size to (0.075m, 0.075m, 0.2m) and set the detection range to [-54m, 54m] in X and Y axes, and [-5m, 3m] in Z axis. We trained SAFDNet for 20 epochs to compare with previous methods. The other training settings strictly follow TransFusion-L~\cite{TransFusion}.

% \vspace*{-4mm}
% \paragraph*{Argoverse2} We set the voxel size to (0.1m, 0.1m, 0.2m) and set the detection range to [-200m, 200m] in X and Y axes, and [-4m, 4m] in Z axis. We trained SAFDNet for 24 epochs to compare with previous methods. For the ablation experiments in Section~\ref{ablation_exps}, we trained all models for 6 epochs.

\subsection{Comparison with state-of-the-art methods}

\paragraph*{Results on Waymo Open.}  On the validation set, SAFDNet achieved 75.7\% L2 mAP and 73.9\% L2 mAPH, performing slightly better than the hybrid detector HEDNet and the sparse detector FSDv2. On the test set, SAFDNet yielded 1.3\% L2 mAPH gains over the previous best fully sparse detector FSDv2. Notably, SAFDNet achieved significant improvements over FSDv2 on the large vehicle category (2.2\% L2 mAPH on the test set), which suffers more from the center feature missing problem with the full sparse architecture. More comparison on model runtime has been presented later.

\vspace*{-4mm}
\paragraph*{Results on nuScenes.} We primarily compared SAFDNet with previous top-performing LiDAR-based methods on the nuScenes test set (Table~\ref{nuscenes_results}). On the nuScenes test set, SAFDNet achieved impressive results with 72.3\% NDS and 68.3\% mAP. Compared with FSDv2, SAFDNet showcased significant improvements on the categories of large objects including car (+3.6\%), truck (+5.7\%), and trailer (+4.6\%). These results further demonstrate the effectiveness of our method.

\begin{table*}[t]
    \begin{center}
    \scalebox{0.77}{
       \setlength{\tabcolsep}{0.65mm}{}
       \hspace{-4mm}
       \begin{tabular}{l|c|rrrrrrrrrrrrrrrrrrrrrrrrrrrrrrrr}
       \thickhline
      Method & \rotatebox{90}{mAP} & \rotatebox{90}{Vehicle} & \rotatebox{90}{Bus} & \rotatebox{90}{Pedestrian} & \rotatebox{90}{Stop Sign} &\rotatebox{90}{Box Truck} &\rotatebox{90}{Bollard} &\rotatebox{90}{C-Barrel} &\rotatebox{90}{Motorcyclist} &\rotatebox{90}{MPC-Sign} &\rotatebox{90}{Motorcycle} &\rotatebox{90}{Bicycle} &\rotatebox{90}{A-Bus} &\rotatebox{90}{School Bus} &\rotatebox{90}{Truck Cab} &\rotatebox{90}{C-Cone} &\rotatebox{90}{V-Trailer} &\rotatebox{90}{Sign} &\rotatebox{90}{Large Vehicle} &\rotatebox{90}{Stroller} &\rotatebox{90}{Bicyclist} &\rotatebox{90}{Truck} & \rotatebox{90}{MBT} &\rotatebox{90}{Dog} & \rotatebox{90}{Wheelchair} & \rotatebox{90}{W-Device}  & \rotatebox{90}{W-Rider} \\
      \midrule
    %   CenterPoint & 13.5 & 61.0 & 36.0 & 33.0 & 28.0 & 26.0 & 25.0 & 22.5 & 16.0 & 16.0 & 12.5 & 9.5 & 8.5 & 7.5 & 8.0 & 8.0 & 7.0 & 6.5 & 3.0 & 2.0 & 14.0 & 14.0 & 1.0 & 0.5 & 0 & 3.0 & 0 \\
      CenterPoint~\cite{CenterPoint} & 22.0 & 67.6 & 38.9 & 46.5 & 16.9 & 37.4 & 40.1 & 32.2 & 28.6 & 27.4 & 33.4 & 24.5 & 8.7 & 25.8 & 22.6 & 29.5 & 22.4 & 6.3 & 3.9 & 0.5 & 20.1 & 22.1 & 0.0 & 3.9 & 0.5 & 10.9 & 4.2 \\
      HEDNet~\cite{HEDNet} & 37.1 & 78.2 & 47.7 & 67.6 & 46.4 & 45.9 & 56.9 & 67.0 & 48.7 & 46.5 & 58.2 & 47.5 & 23.3 & 40.9 & 27.5 & 46.8 & 27.9 & 20.6 & 6.9 & 27.2 & 38.7 & 21.6 & 0.0 & 30.7 & 9.5 & 28.5 & 8.7 \\
      \midrule
      VoxelNeXt~\cite{VoxelNeXt}$^\dag$ & 30.7 & 72.7 & 38.8 & 63.2 & 40.2 & 40.1 & 53.9 & 64.9 & 44.7 & 39.4 & 42.4 & 40.6 & 20.1 & 25.2 & 19.9 & 44.9 & 20.9 & 14.9 & 6.8 & 15.7 & 32.4 & 16.9 & 0.0 & 14.4 & 0.1 & 17.4 & 6.6 \\
      FSDv1~\cite{FSD}$^\dag$ & 28.2 & 68.1 & 40.9 & 59.0 & 29.0 & 38.5 & 41.8 & 42.6 & 39.7 & 26.2 & 49.0 & 38.6 & 20.4 & 30.5 & 14.8 & 41.2 & 26.9 & 11.9 & 5.9 & 13.8 & 33.4 & 21.1 & 0.0 & 9.5 & 7.1 & 14.0 & 9.2 \\
      FSDv2~\cite{FSDv2}$^\dag$ & 37.6 & 77.0 & 47.6 & 70.5 & 43.6 & 41.5 & 53.9 & 58.5 & 56.8 & 39.0 & 60.7 & 49.4 & 28.4 & 41.9 & 30.2 & 44.9 & 33.4 & 16.6 & 7.3 & 32.5 & 45.9 & 24.0 & 1.0 & 12.6 & 17.1 & 26.3 & 17.2 \\
      \rowcolor{cyan! 20} SAFDNet (Ours)$^\dag$ & \textbf{39.7} & 78.5 & 49.4 & 70.7 & 51.5 & 44.7 & 65.7 & 72.3 & 54.3 & 49.7 & 60.8 & 50.0 & 31.3 & 44.9 & 24.7 & 55.4 & 31.4 & 22.1 & 7.1 & 31.1 & 42.7 & 23.6 & 0.0 & 26.1 & 1.4 & 30.2 & 11.5 &  \\
       \thickhline
       \end{tabular}}
    \end{center}
    \vspace*{-5mm}
    \caption{Comparison with prior methods on Argoverse2 validation set. Metrics: mAP (\%)$\uparrow$ for the overall results, AP (\%)$\uparrow$ for each category.}
    \label{argo2_results}
    \vspace*{-3mm}
\end{table*}

\vspace*{-4mm}
\paragraph*{Results on Argoverse2.} To validate the effectiveness of SAFDNet on the long-range detection, we conducted experiments on the Argoverse2 dataset with a perception range of 200 meters (Table~\ref{argo2_results}). SAFDNet achieved a gain of 2.1\% mAP over the previous best sparse detector FSDv2. It is worth noting that the proposed SAFDNet also outperformed the hybrid detector HEDNet, which indicates that expanding features towards all unoccupied area without restriction may hurt the detection performance.

\begin{table}[t]
    \centering
    \hspace*{-2mm}
    \scalebox{0.88}{
        \setlength{\tabcolsep}{0.5mm}
        \begin{tabular}{l|ccc|cccc}
        \toprule
        \multirow{2}{*}{Method} & \multicolumn{3}{c|}{Waymo Open} & \multicolumn{4}{c}{Argoverse2} \\
        & mAPH & FPS & \small{Speedup} & mAP & FPS & \small{Speedup} & Mem. \\
        \midrule
        HEDNet~\cite{HEDNet}  & 73.4 & 17.2 & 1.0$\times$ & 37.1 & 7.3 & 1.0$\times$ & 28.7G  \\
        \midrule
        VoxelNeXt~\cite{VoxelNeXt}$^\dag$ & 70.1 & 15.7 & 0.9$\times$ & 30.7 & 19.6 & 2.7$\times$& 6.2G \\
        FSDv2~\cite{FSDv2}$^\dag$ & 73.5 & 10.3 & \underline{0.6$\times$} & \underline{37.6} & 11.5 & 1.6$\times$ & 8.6G \\
        \rowcolor{cyan! 20} SAFDNet$^\dag$(Ours) & 73.9 & 20.2 & \underline{1.2$\times$} & \underline{39.7} & 15.1 & 2.1$\times$ & 7.3G \\
        \bottomrule
        \end{tabular}
    }
    \vspace*{-2mm}
    \caption{Runtime comparison on the Waymo Open and Argoverse2 datasets. Mem. denotes the training GPU memory measured following~\cite{MMDetection3D}. FPS (frame per second, $\uparrow$) is the inference speed measured on a single NVIDIA 4090 GPU with a batch size of 1.}
    \label{runtime}
    % \vspace*{-1mm}
\end{table}

\vspace*{-4mm}
\paragraph*{Runtime comparison.} We compared the inference speed of SAFDNet with prior top-performing methods, as shown in Table~\ref{runtime}. On the Waymo Open dataset with a short perception range, SAFDNet was 2$\times$ faster than the sparse detector FSDv2 and was 1.2$\times$ faster than the hybrid detector HEDNet. On the Argoverse2 dataset with a long perception range, SAFDNet yielded 2.1\% mAP gains over FSDv2 while being 1.3$\times$ faster. Compared with HEDNet, SAFDNet achieved 2.6\% mAP improvements while being 2.1$\times$ faster. Note that the hybrid detector HEDNet requires much more training memory than the other sparse detectors.

\begin{table}[t]
    \begin{minipage}[t]{1.0\linewidth}
        \centering
        \scalebox{0.92}{
            \setlength{\tabcolsep}{0.7mm}{}
            \begin{tabular}{c|c|cccc}
            \toprule
            No. & Method & mAPH & Veh. & Ped. & Cyc. \\
            \midrule
            * & HEDNet & 73.2 & 72.1 & 72.0 & 75.6 \\
            \midrule
            1 & Center heatmap &  55.2 & 53.3 & 54.5 & 60.7 \\
            2 & Nearest heatmap & 71.0 & 69.1 & 70.6 & 73.3 \\
            3 & Normalized heatmap & 71.4 & 69.1 & 70.6 & 74.2 \\
            4 & Row-3 w/ 2D-EDB & 71.5 & 68.8 & 70.9 & 74.7 \\
            \rowcolor{gray! 20} 5 & Row-4 w/ AFD & 73.3 & 71.7 & 72.3 & 75.7 \\
            \bottomrule
            \end{tabular}
        }
        \vspace*{-2mm}
        \caption*{(a) A step-by-step ablation.}
        \vspace*{2mm}
    \end{minipage}

    \begin{minipage}[t]{1.0\linewidth}
        \centering
        \scalebox{0.95}{
            \setlength{\tabcolsep}{2.0mm}
            \begin{tabular}{c|cc|cc}
            \toprule
            \multirow{2}{*}{Type} & \multicolumn{2}{c|}{Waymo Open} & \multicolumn{2}{c}{Argoverse2} \\
            & mAPH & FLOPs & mAP & FLOPs \\
            \midrule
            w/o  & 71.5 & 90G & 36.4 & 34G\\
            PB UFD & 73.0 & 189G & 37.6 & 147G \\
            PF UFD & 73.1 & 182G  & 37.7 & 144G\\
            \rowcolor{gray! 20} AFD & 73.3 & 108G & 37.8 & 45G \\
            \bottomrule
            \end{tabular}
        }
        \vspace*{-2mm}
        \caption*{(b) Different types of feature diffusion.}
        \vspace*{2mm}
      \end{minipage}

    \begin{minipage}[t]{1.0\linewidth}
       \centering
       \scalebox{0.95}{
        \setlength{\tabcolsep}{2.8mm}
        \begin{tabular}{c|cc|cc}
        \toprule
            \multirow{2}{*}{$\alpha$} & \multicolumn{2}{c|}{Waymo Open} & \multicolumn{2}{c}{Argoverse2} \\
            & mAPH & FLOPs & mAP & FLOPs \\
            \midrule
            0.0 &  71.5 & 90G & 36.4 & 34G\\
            0.5 &  72.7 & 100G & 37.3 & 38G \\
            \rowcolor{gray! 20} 1.0 & 73.3 & 108G & 37.8 & 45G \\
            2.0 &  73.0 & 153G & 37.8 & 67G \\
            \bottomrule
        \end{tabular}
        }
        \vspace*{-2mm}
       \caption*{(c) Different ranges of adaptive feature diffusion.}
    \end{minipage}

    \vspace*{-2mm}
    \caption{Ablation studies. The overall and per-category L2 mAPH on Waymo Open and the mAP on Argoverse2 are presented.  The hyperparameter $\alpha$ controls the feature diffusion range. FLOPs were calculated excluding the 3D sparse backbone. Veh., Ped. and Cyc. are short for vehicle, pedestrian, and cyclist, respectively.}
    \label{ablations}
    \vspace*{-2mm}
\end{table}

\subsection{Ablation studies}\label{ablation_exps}
% We conducted ablation experiments on the Waymo Open and Argoverse2 datasets to better understand SAFDNet.

\begin{figure*}[t]
    \centering
    \includegraphics[width=16cm]{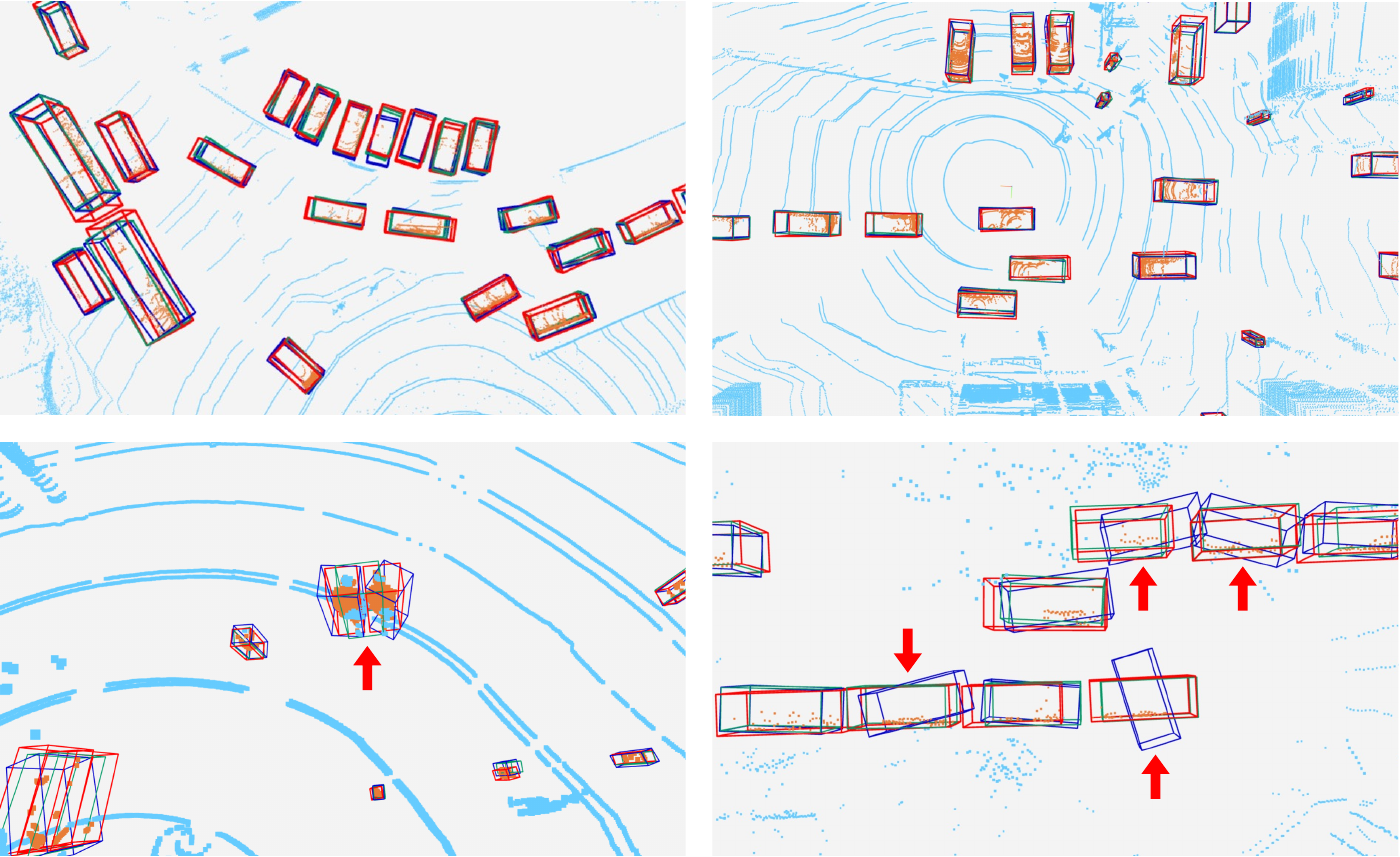}
    \caption{Qualitative results on Argoverse2. The red, blue, and green boxes are human annotations, SAFDNet predictions and HEDNet predictions, respectively. The orange points denote the points that fall within the human-annotated boxes. SAFDNet performed comparably to HEDNet in some scenarios (top row). Additionally, SAFDNet demonstrated better predictions for small objects (bottom-left panel) but encountered challenges in direction prediction for \textit{partially} large objects (bottom-right panel). Red arrows mark the prediction differences.}
    \label{visualization}
\end{figure*}

% \vspace*{-4mm}
\paragraph*{A step-by-step ablation.} We performed step-wise ablation experiments from the hybrid detector HEDNet to our SAFDNet on the Waymo Open dataset (Table~\ref{ablations} (a)). Initially, we removed the 2D dense backbone in HEDNet and replaced all convolutions in the detection head with submanifold sparse convolutions, resulting in the first model. The accuracies across all categories dropped significantly. We observed that using the voxels within which object centers fall, as done in CenterPoint, to calculate the Gaussian heatmap during classification training led to rapid convergence of the classification loss to zero. In the second model, we used the nearest non-empty voxel as the center to generate the Gaussian heatmap. This adjustment notably increased accuracy to 71.0\% mAPH. In the third model, we continued to adopt the center voxel but normalized the training target with the maximum values to prevent gradient vanishing, resulting in a slight accuracy boost. Adding the 2D-EDB in the fourth model did not lead to a performance improvement. Notably, there still existed a significant accuracy gap, particularly on large vehicles, between the fourth model and HEDNet (over 3\% mAPH). Finally, incorporating the proposed adaptive feature diffusion strategy effectively bridged this gap. An ablation on different sparse backbones is presented in \textbf{Appendix B}.

\vspace*{-4mm}
\paragraph*{Different types of feature diffusion.} We compared different type of diffusion strategies on the Waymo Open and Argoverse2 datasets, as presented in \mbox{Table~\ref{ablations} (b)}. While the models with the two types of UFD strategies showed substantial gains over the model without feature diffusion, they introduced significantly higher computational FLOPs. In contrast, the model with the AFD strategy achieved the highest accuracy while incurring fewer computational costs.

\vspace*{-4mm}
\paragraph*{Different ranges of feature diffusion.} We compared different diffusion ranges of AFD on the Waymo Open and Argoverse2 datasets. Table~\ref{ablations} (c) shows that, setting $\alpha$ to 1, wherein the diffusion kernel size matches the average object size, resulted in the highest accuracy for SAFDNet. On the Waymo Open dataset, a large diffusion range slightly degraded the performance, suggesting that excessive feature diffusion might be unnecessary and could harm performance.

\subsection{Qualitative visualization}
We showcase predictions made by HEDNet and SAFDNet on the Argoverse2 dataset in Figure~\ref{visualization}. SAFDNet performed comparably to HEDNet in some scenarios, exhibiting better predictions for small objects but encountering challenges in predicting the direction of certain large objects. Please note that this issue arises only with partially large objects, like box truck, and is not a general problem for all large objects.

\section{Conclusion}

We present SAFDNet, a fully sparse architecture tailed for 3D object detection. To address the center feature missing problem, we propose an adaptive feature diffusion strategy to diffuse features towards object centers while maintaining feature sparsity as much as possible. SAFDNet achieved impressive performance on the Waymo Open, nuScenes, and Argoverse2 datasets, which demonstrates the effectiveness of our method. We hope that our work can provide some inspiration for the design of fully sparse 3D object detector.

\vspace*{-4mm}
\paragraph*{Limitations.} We mitigate the center feature missing problem through our proposed adaptive feature diffusion strategy. However, this approach may generate some unnecessary feature regions, such as the expanded regions outside of objects. We believe a more efficient solution could address this issue, such as by grouping voxels associated with the same object. We defer exploration of this solution to future research.

\vspace*{-4mm}
\paragraph*{Acknowledgements.}

This work was supported by the National Key Research and Development Program of China (grant 2021ZD0200301), the National Natural Science Foundation of China (grant U2341228), and the Shanghai Automotive Industry Corporation (SAIC) Intelligent Technology (contract CGHT-202112218).

{
    \small
    \bibliographystyle{unsrt}
    \bibliography{safdnet}
}

\newpage
{\color{white}.}

\newpage

\section*{Appendix A Implementations details}
We trained all models with a batch size of 16, employed the Adam~\cite{Adam} optimizer with a one-cycle learning rate policy, and set the weight decay to 0.05 and the maximum learning rate to 0.003. More details are presented below.

\vspace*{-4mm}
\paragraph*{Waymo Open dataset.} We adopt a detection head derived from CenterPoint. We set the voxel size to (0.08m, 0.08m, 0.15m), and the detection range to [-75.52m, 75.52m] in X and Y axes, and [-2m, 4m] in Z axis. We trained SAFDNet for 24 epochs on the training dataset and reported the evaluation results on the validation set to compare with previous methods. For results on the test set, we trained the model on both the training and validation sets. We adopted the faded training strategy in the last epoch. During inference, we applied class-specific NMS with an IoU threshold of 0.75, 0.6 and 0.55 for vehicle, pedestrian, and cyclist, respectively. There are three category groups: group 1 includes vehicle; group 2 includes pedestrian and cyclist; group 3 is the background. The kernel size K for feature diffusion for the three groups are set to 7, 3, and 3, respectively.

\vspace*{-4mm}
\paragraph*{nuScenes dataset.} We adopt a detection head derived from TransFusion-L.We set the voxel size to (0.3m, 0.3m, 8.0m), and the detection range to [-54m, 54m] in X and Y axes, and [-5m, 3m] in Z axis. We trained SAFDNet for 20 epochs on both training and validation sets and reported the evaluation results on the test set to compare with previous methods. The faded strategy was used during the last 4 epochs. We set the query number of detection head to 300 (test only) and did not use any test-time augmentation. There are four category groups: group 1 includes bus and trailer; group 2 includes car, truck, and construction vehicle; group 3 includes motorcycle, bicycle, traffic cone, pedestrian, and barrier; group 4 is the background. The kernel size K for feature diffusion for the four groups are set to 15, 9, 7, and 7, respectively.

\vspace*{-4mm}
\paragraph*{Argoverse2 dataset.}
We adopt a detection head derived from CenterPoint. We set the voxel size to (0.1m, 0.1m, 0.2m), and the detection range to [-200m, 200m] in X and Y axes, and [-4m, 4m] in Z axis. We trained SAFDNet for 24 epochs on the training set and reported the results on the validation set to compare with other methods. We adopted the faded training strategy in the last epoch. There are four category groups: group 1 includes large vehicle, bus, box truck, truck, truck cab, vehicular trailer, school bus, articulated bus, and message board trailer; group 2 includes regular vehicle; group 3 includes the other object categories; group 4 is the background. The kernel size K for feature diffusion for the four groups are set to 13, 7, 3, and 3, respectively.

\section*{Appendix B More experimental results}

\begin{table}[t]
    \centering
    \hspace*{-2mm}
    \scalebox{0.9}{
        \setlength{\tabcolsep}{1.0mm}{}
        \begin{tabular}{ccc|cccc}
        \toprule
        3D Backbone & Other Parts & AFD & mAPH & Veh. & Ped. & Cyc. \\
        \midrule
        HEDNet & Dense  & & 73.2 & 72.1 & 72.0 & 75.6 \\
        HEDNet & Sparse  & & 71.5 & 68.8 & 70.9 & 74.7 \\
        \rowcolor{gray! 20} HEDNet & Sparse & \checkmark & \textbf{73.3} & 71.7 & 72.3 & 75.7 \\
        \midrule
        VoxelNet & Dense  & & 71.4 & 69.8 & 70.9 & 73.4 \\
        VoxelNet & Sparse  & & 69.9 & 66.1 & 70.6 & 72.9 \\
        \rowcolor{gray! 20} VoxelNet & Sparse & \checkmark & \textbf{72.0} & 70.2 & 71.7 & 74.1 \\
        \midrule
        PillarNet & Dense  & & 68.7 & 69.2 & 66.5 & 70.5 \\
        PillarNet & Sparse  & & 67.2 & 65.6 & 66.8 & 70.2 \\
        \rowcolor{gray! 20} PillarNet & Sparse & \checkmark & \textbf{69.4} & 69.7 & 67.5 & 71.4 \\
        \bottomrule
        \end{tabular}
    }
    \caption{Adaptive feature diffusion(AFD) on different backbones.}
    \label{tab:different_backbones}
\end{table}

We conducted experiments with three 3D sparse backbones: HEDNet, VoxelNet, and PillarNet. Table~\ref{tab:different_backbones} shows that the proposed AFD module worked well on all three backbones.

\end{document}

%% file: preamble.tex
\usepackage[dvipsnames]{xcolor}